\documentclass[acmtog,nonacm]{acmart}

\usepackage{multirow}
\usepackage[ruled,vlined,linesnumbered]{algorithm2e}

\AtBeginDocument{%
  }

\newcommand{\method}{SVG360}

\title{\method: Editable Multiview Vector Graphics from a Single SVG}

\author{Mengnan Jiang}
\affiliation{%
  \institution{Mercedes-Benz AG}
  \country{Germany}}
\affiliation{%
  \institution{Technical University of Darmstadt}
  \country{Germany}}
\email{mengnan.jiang@mercedes-benz.com}

\author{Zhaolin Sun}
\affiliation{%
  \institution{Technical University of Darmstadt}
  \country{Germany}}

\author{Christian Franke}
\affiliation{%
  \institution{Mercedes-Benz AG}
  \country{Germany}}
\email{christian.franke@mercedes-benz.com}

\author{Michele Franco Adesso}
\affiliation{%
  \institution{Mercedes-Benz AG}
  \country{Germany}}
\email{michele\_franco.adesso@mercedes-benz.com}

\author{Antonio Haas}
\affiliation{%
  \institution{Mercedes-Benz AG}
  \country{Germany}}
\affiliation{%
  \institution{University of Stuttgart}
  \country{Germany}}
\email{antonio.haas@mercedes-benz.com}

\author{Grace Li Zhang}
\affiliation{%
  \institution{Technical University of Darmstadt}
  \country{Germany}}
\email{grace.zhang@tu-darmstadt.de}

\renewcommand{\shortauthors}{Jiang et al.}

\makeatletter
\@ACM@balancefalse
\makeatother

\begin{document}

\begin{teaserfigure}
    \centering
    \includegraphics[width=0.85\textwidth]{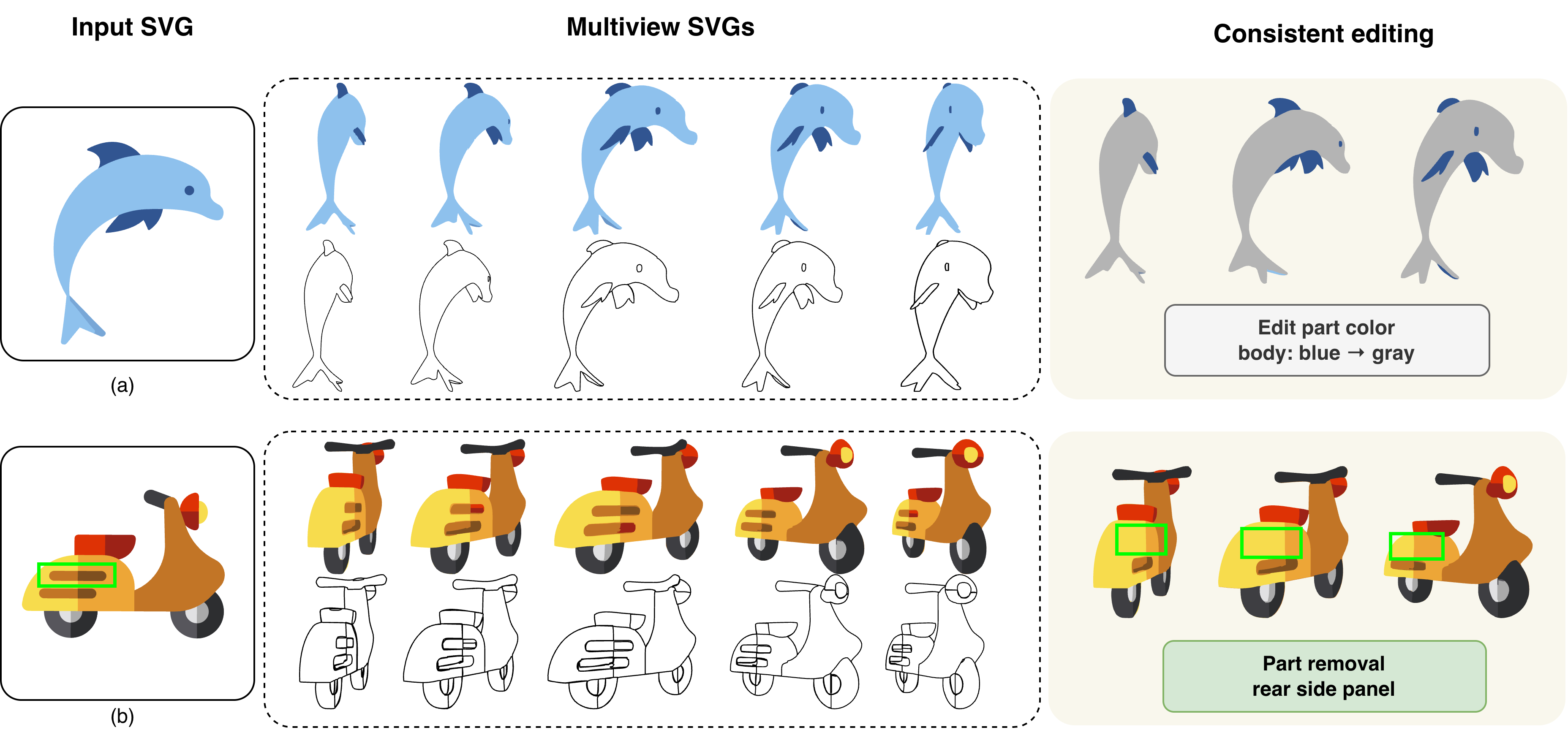}
    \caption{\textbf{Examples of editable multiview SVG asset generation.}
  Given a single input SVG, \method{} generates coherent multiview SVGs together with their vector path visualizations.
  The examples show that the output assets preserve cross-view consistency and remain easy to edit: the dolphin body color can be changed consistently across views, and the scooter rear side panel can be removed as a selected part.}
    \Description{The figure shows two examples of SVG360. Each example contains an input SVG, several generated multiview SVG outputs with corresponding path visualizations, and a consistent editing example. The dolphin example changes the body color across views, while the scooter example removes the rear side panel.}
    \label{fig:teaser}
\end{teaserfigure}

\begin{abstract}
Scalable Vector Graphics are a standard representation for editable visual design, yet they are usually authored as single view two dimensional illustrations. This limits their use in applications that require object level assets to remain coherent when observed, edited, or animated from different viewpoints. We present SVG360, a framework that converts a single input SVG into geometrically and visually consistent multiview SVG assets. The key challenge is that direct per view generation or vectorization produces view dependent regions, fragmented paths, and unstable colors, making the resulting SVGs difficult to edit as a coherent object. SVG360 addresses this problem through a view consistent vectorization pipeline. It first lifts the rasterized input into a view conditioned object representation and renders target views under prescribed cameras. It then propagates part identity across neighboring views using a spatial memory mechanism adapted from video segmentation, establishing consistent region decomposition, path correspondence, and color assignment without task specific retraining. Finally, each view is reconstructed as an editable SVG through structure aware vectorization, where redundant paths are consolidated and local geometry is optimized while preserving boundaries and semantic parts. Experiments on object level SVG assets show that SVG360 improves multiview consistency, reduces path redundancy, and better preserves fine structures compared with direct per view vectorization. By turning a single view SVG into a coherent 360 degree vector asset, SVG360 expands vector graphics from static illustration toward editable multiview content for design, animation, and structured visual editing.
\end{abstract}

\maketitle

\section{Introduction}
\label{sec:intro}

Editable vector graphics are central to modern design workflows because they provide resolution independence, compact structure, and precise editability. 
However, most SVG assets are authored as single-view two-dimensional illustrations. 
When designers need to present an existing vector artwork from a new viewpoint, they often have to manually redraw or adjust its paths, which is time consuming and can introduce geometric, stylistic, and color inconsistencies. 
Adobe Illustrator's Turntable (Beta)~\cite{adobeTurntableHelp} is one of the few publicly available tools for generating multiple views from a single 2D artwork, but it provides limited viewpoint control and discloses no technical details.

We study the problem of generating editable multiview SVG assets from a single input SVG. 
The goal is not only to synthesize plausible novel views, but also to produce a coherent family of path-level editable SVGs that preserve the input design style, maintain stable part structure, and remain compact across viewpoints. 
This task is valuable for turntable visualization, multiview icon and logo design, animation asset preparation, and structured visual editing. 
Yet it remains underexplored because it lies between generative view synthesis, which usually produces raster images, and vector graphics reconstruction, which usually focuses on single-view SVG recovery.
Figure~\ref{fig:teaser} illustrates our goal: converting one input SVG into consistent multiview SVGs that expose editable paths and support part-level editing across viewpoints.

Recent generative models have enabled impressive single-image novel view synthesis and multiview image generation~\cite{ho2020ddpm,rombach2022ldm,nichol2021glide,saharia2022imagen,ramesh2022dalle2,zhang2023controlnet,mou2023t2iadapter,liu2023zero123,shi2023zero123pp,liu2023syncdreamer,voleti2024sv3d}. 
However, their outputs are raster-based and do not expose editable paths, palette structure, or part-level organization. 
Conversely, raster-to-vector and SVG generation methods such as DiffVG~\cite{li2020diffvg}, DeepSVG~\cite{carlier2020deepsvg}, Im2Vec~\cite{reddy2021im2vec}, LIVE~\cite{yue2024live}, and StarVector~\cite{wang2024starvector} focus mainly on single-view reconstruction. 
Applying them independently to each generated view often produces fragmented paths, inconsistent boundaries, unstable palettes, and view-dependent topology, so the resulting SVGs may look plausible individually but fail to behave as a coherent editable object.

To address this challenge, we propose \method{}, a framework that converts a single input SVG into geometrically and visually consistent multiview SVG assets. 
First, the input SVG is rasterized and lifted into a 3D-aware representation to render plausible target views under prescribed cameras. 
Second, we introduce a spatially aware cross-view part propagation module that adapts video segmentation memory to the viewing sphere. 
Instead of using temporal adjacency, it selects geometrically nearby views as memory references and propagates part identity across viewpoints, reducing view-dependent region drift before vectorization. 
Third, we perform part-aware raster-to-vector conversion followed by SVG refinement, including path cleanup and palette alignment, suppressing redundant paths while preserving semantic boundaries and stable colors.

Experiments show that \method{} produces cleaner and more stable multiview SVGs than Adobe Illustrator's Turntable. 
It reduces path-count deviation $\mathrm{RMSE}_{\text{path}}$, lowers adjacent-view color-count variation $\overline{\Delta N}_{\text{color,nbr}}$, and reduces perceptual color drift across views. 
It also improves overall multiview continuity measured by DINO feature consistency and DreamSim perceptual similarity~\cite{oquab2023dinov2,fu2023dreamsim}. 
Beyond Turntable, we further compare rendered multiview results with related view synthesis and 3D generation methods to evaluate general multiview consistency, and ablations confirm the importance of spatial part propagation and SVG refinement.

Our contributions are summarized as follows.

\noindent\textbf{Editable Multiview SVG Asset Generation.}
We formulate the task of converting a single-view SVG into a coherent family of editable multiview SVGs, bridging generative view synthesis and structured vector graphics.

\noindent\textbf{Spatially Aware Cross-View Part Propagation.}
We introduce a spatial memory mechanism that replaces temporal adjacency with geometry-guided neighborhood traversal on the viewing sphere, improving part-level consistency across generated views.

\noindent\textbf{Structure-Preserving SVG Refinement.}
We develop an SVG refinement strategy that removes redundant paths, aligns colors to a stable palette, and preserves semantic boundaries, producing compact SVG assets suitable for downstream design workflows.


\section{Related Work}
\label{sec:related_work}

\subsection{Single-Image to 3D Vector Graphics}
\label{sec:3d_vector}

Recent work has explored generating or reconstructing vector-like 3D representations from a single image or text prompt~\cite{choi2024_3doodle,li2025_empowering,wang2025_viewcraft3d}. 
These methods use 3D curves, vector primitives, or hybrid representations to support novel-view rendering with stronger geometric consistency than independent 2D generation. 
However, they usually aim to create a new 3D vector representation rather than preserve and extend an existing SVG asset. 
In contrast, our goal is to generate editable \emph{2D} SVG views from a given input SVG while preserving its visual identity, region structure, palette behavior, and path-level editability.

\begin{figure*}[t]
  \centering
  \includegraphics[width=0.87\textwidth]{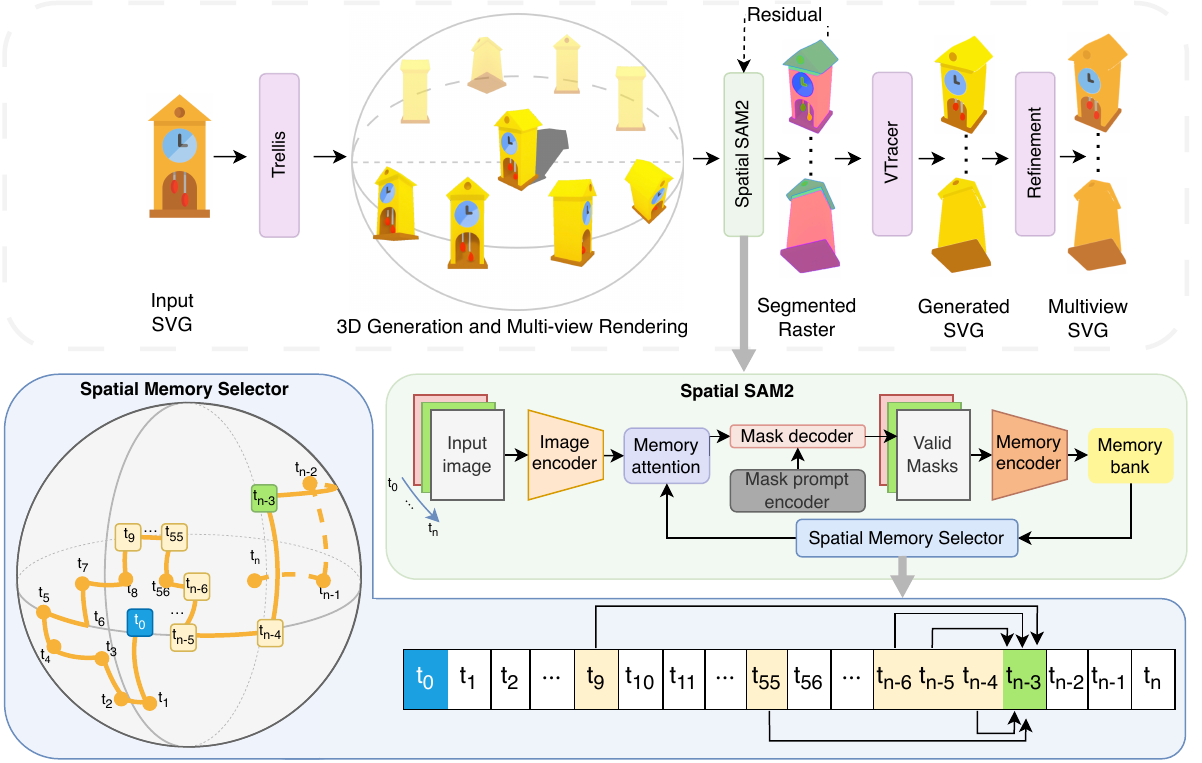}
   \caption{Overview of \method{}. Given a single input SVG, our pipeline first rasterizes the artwork and lifts it into a 3D-aware representation to render multiview rasters. Spatial SAM2 propagates part masks across the viewing sphere by selecting angularly nearby views as memory support, with a lightweight residual discovery step to recover newly visible regions. The segmented rasters are then converted into SVGs by part-wise VTracer vectorization and further improved through SVG refinement, including path cleanup and palette alignment. The final output is a coherent family of compact, editable, and cross-view consistent multiview SVG assets.}
  
  \Description{The figure shows the SVG360 pipeline. An input SVG is rasterized and used for 3D generation and multiview rendering. The rendered views are processed by Spatial SAM2, which uses a spatial memory selector on the viewing sphere to propagate consistent part masks. A residual discovery step indicates recovery of missing regions. The segmented rasters are vectorized with VTracer and then refined through path cleanup and palette alignment to produce multiview SVG outputs.}
  \label{fig:main_pipeline}
\end{figure*}


\subsection{Single-Image to Multi-View Generation}
\label{sec:single_view_to_multi_view}

Single-image multiview generation is a natural precursor to editable multiview SVG generation. 
Image-domain methods, such as Zero-1-to-3, Zero123++~\cite{liu2023zero123,shi2023zero123pp}, SyncDreamer~\cite{liu2023syncdreamer}, and SV3D~\cite{voleti2024sv3d}, synthesize target views directly as rasters. 
They can produce plausible novel views, but do not expose editable paths, explicit region structure, or stable palette assignments, and may still suffer from shape drift or inconsistent details across viewpoints.

Another line of work infers a shared 3D asset and renders novel views from the reconstructed geometry. 
Representative methods include mesh generators such as GET3D, PolyGen, and MeshDiffusion~\cite{gao2022get3d,polygen,meshdiffusion}, single-image reconstruction and multiview generation models such as Wonder3D~\cite{long2024wonder3d}, LRM, TripoSR, and Unique3D~\cite{hong2023lrm,triposr,wu2024unique3d}, and latent 3D models such as Shape-E and Trellis~\cite{shapee,trellis2024}. 
Similarly, 3D Gaussian Splatting~\cite{kerbl2023gaussians} and related generative extensions~\cite{tang2023dreamgaussian,gaussiandreamer2023,splatterimage2024} provide efficient view-consistent raster rendering from a shared 3D field. 
These methods improve geometric consistency, but their outputs remain raster renderings or 3D assets rather than editable SVGs. 
They also do not provide compact vector topology, cross-view path correspondence, or palette-stable SVG structure. 
Thus, they are complementary to multiview SVG asset generation rather than direct substitutes.

\subsection{Raster-to-Vector Generation}
\label{sec:raster_to_vector}

Converting raster images into structured vector graphics is a long-standing problem. 
Classical tracing methods such as Potrace~\cite{selinger2003potrace} and AutoTrace~\cite{weber2001autotrace} work well for simple icons and low-color graphics, but often produce fragmented paths under antialiased boundaries, smooth shading, or noisy color transitions.
VTracer~\cite{vtracer2023} extends practical tracing to color images through color layering and spline fitting.

Optimization-based methods such as DiffVG~\cite{li2020diffvg} optimize vector parameters through differentiable rendering, while learning-based approaches such as DeepSVG~\cite{carlier2020deepsvg} and Im2Vec~\cite{reddy2021im2vec} model vector graphics using neural sequence or latent representations. 
Recent methods further improve topology, compactness, or path quality. 
For example, LIVE~\cite{yue2024live} uses layer-wise vectorization to preserve global structure, while VectorFusion~\cite{jain2022vectorfusion} and StarVector~\cite{wang2024starvector} introduce generative or region-aware priors for SVG reconstruction.

Most existing raster-to-vector methods focus on single-view reconstruction. 
Applying them independently to multiple generated views can cause unstable path counts, inconsistent boundaries, palette drift, and view-dependent topology. 
For multiview SVG generation, vectorization should therefore be guided by cross-view part structure rather than only local pixel transitions. 
Our method addresses this by performing part-aware vectorization after spatially consistent segmentation, producing cleaner paths and more stable SVG assets across viewpoints.

\subsection{Segmentation and Cross-View Consistency}
\label{sec:segmentation_consistency}

Segmentation can simplify vectorization by decomposing an image into coherent regions before tracing. 
However, multiview SVG generation requires the same semantic part to remain consistently identified across viewpoints. 
Without such consistency, independently vectorized views may assign different regions, merge nearby parts, or lose small structures, making the SVG sequence difficult to edit as one coherent asset.

Conventional segmentation methods~\cite{he2017maskrcnn,cheng2022mask2former,xie2021segformer,li2023maskdino} focus on per-image accuracy and do not enforce label correspondence across views. 
Foundation segmentation models such as SAM, HQ-SAM, and SAM2~\cite{kirillov2023sam,cheng2023hqsam,ravi2024sam2} provide strong boundary quality and flexible prompting, but their propagation mechanisms are mainly designed for temporal video sequences. 
Direct temporal tracking can fail when neighboring indices are not geometrically adjacent on the viewing sphere.

Memory-based video object segmentation methods~\cite{cheng2021stcn,yang2023deAOT} maintain masks across time by storing and retrieving features from previous frames. 
We adapt this idea to multiview SVG generation by treating cameras as samples on a viewing sphere and retrieving spatially relevant views by angular proximity. 
This bridges cross-view part consistency and vector-level SVG reconstruction, enabling compact, palette-stable, and editable SVG assets across viewpoints.

\section{Method}
\label{sec:method}

As illustrated in Figure~\ref{fig:main_pipeline}, our \textsc{SVG360} framework converts a single input SVG into a coherent family of editable multiview SVG assets.
The pipeline first rasterizes the input SVG and lifts it into a 3D-aware representation to render geometrically plausible target views (\S\ref{sec:multi_view_raster_generation}).
It then performs \textbf{spatially aligned segmentation propagation} (\S\ref{sec:segmentation}) to establish consistent part-level regions across viewpoints, with a lightweight residual discovery step for newly visible regions.
Finally, part-wise vectorization followed by SVG refinement (\S\ref{sec:vector_domain_refinement}) converts the segmented rasters into compact, editable multiview SVGs with stable path structure and palette behavior.

\subsection{Multi-View Raster Generation}
\label{sec:multi_view_raster_generation}

Generating accurate and geometrically consistent multiview images from a single SVG input is challenging, since purely 2D generation approaches often struggle to maintain structural coherence across viewpoints.
To obtain a stable raster support for subsequent segmentation and vectorization, we adopt a generative 3D-based approach to synthesize target views.

Although several recent single-image 3D methods, such as Sparc3D~\cite{sparc3d2025} and Hi3DGen~\cite{hi3dgen2025}, demonstrate strong geometric fidelity, they primarily focus on precise shape reconstruction and often do not provide fully textured, render-ready outputs.
Such geometry-centered representations are less suitable for our color-consistent raster-to-SVG workflow, where each generated view must provide both object structure and appearance cues for vector reconstruction.
Considering open-source availability, generation efficiency, and the level of visual fidelity required by our task, we adopt Trellis~\cite{trellis2024} as the backbone 3D generator.
Trellis offers a practical trade-off between geometric plausibility and texture consistency, making it suitable for producing coherent multiview rasters from a single SVG input.

Trellis represents assets in a unified structured latent space and decodes them into multiple 3D formats, including meshes and 3D Gaussians~\cite{trellis2024}.
Its mesh decoder builds on FlexiCubes~\cite{flexicubes2023}, predicting per-voxel signed-distance values and extracting a watertight surface from the zero-level isosurface.
While mesh decoding provides explicit geometry, converting the latent representation to a mesh and then rasterizing it from many viewpoints introduces additional discretization and baking steps.
These steps may accumulate appearance deviations relative to the input SVG style and incur nontrivial runtime overhead.
For efficiency and appearance fidelity, we directly use the 3D Gaussian decoder and render dense multiview rasters through Gaussian splatting.
The resulting views provide geometrically coherent image observations for cross-view part propagation and part-aware vectorization.

\begin{figure}[htbp]
  \centering
  \includegraphics[width=0.99\linewidth]{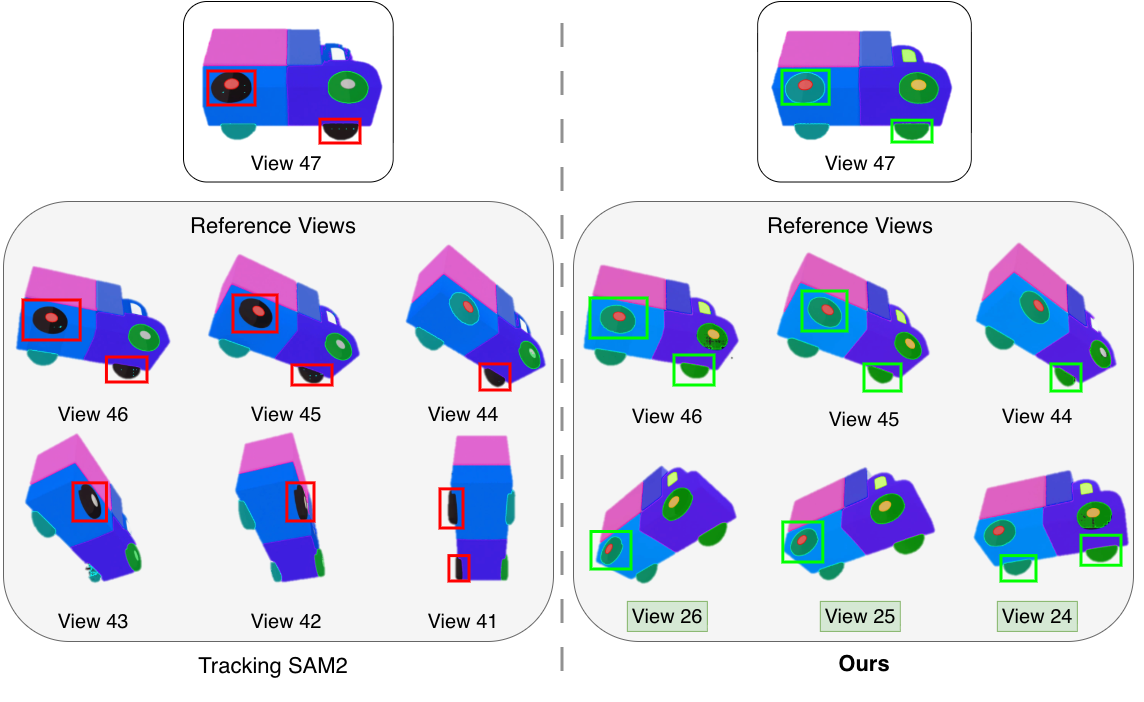}
    \caption{Temporal tracking versus spatial memory selection. 
    For target view~47, SAM2 tracking mode uses temporally adjacent views~41--46 as memory support, while Spatial SAM2 retrieves geometrically nearby views on the viewing sphere, including views~44--46 and views~24--26. 
    Because sequence order does not always match spatial proximity, tracking accumulates errors on small structures such as wheels and lights, whereas spatially selected references provide more reliable part support and reduce mask errors.}
    \Description{A comparison between SAM2 tracking mode and Spatial SAM2 for target view 47. The left side shows SAM2 tracking mode using temporally adjacent reference views 41 to 46, where wheel and light masks become unstable. The right side shows Spatial SAM2 selecting geometrically nearby reference views, including views 44 to 46 and 24 to 26, which provides more reliable support for small parts.}
  \label{fig:spsam2}
\end{figure}

\subsection{Spatially Aligned Segmentation Propagation}
\label{sec:segmentation}

Given the rendered multiview rasters, we need part-level regions that remain consistent across viewpoints.
Independent single-view segmentation cannot enforce label or boundary correspondence, which often leads to unstable vector paths after tracing.
We therefore adapt the memory-based propagation mechanism of SAM2 from temporal video sequences to spatially organized view sequences.
Instead of using temporal adjacency, we define neighborhoods by angular proximity on the viewing sphere and propagate masks through geometrically nearby views.
This encourages each target view to receive memory support from views with similar object visibility, improving part identity and boundary consistency across the full view set.

\noindent\textbf{Spatially sequential scheduling.}

We sample camera viewpoints on a unit sphere and process them using a pseudo-sequential traversal.
Starting from the front view aligned with the input SVG, denoted as $\theta_0$, each subsequent view is chosen as the unvisited view with the smallest angular distance from the current one:
\begin{equation}
\theta_{t+1} = \arg\min_{\theta \in \mathrm{unvisited}} d(\theta_t, \theta).
\end{equation}
This nearest-neighbor traversal reduces abrupt viewpoint changes during mask propagation.
For two viewpoints $\theta_i$ and $\theta_j$, we compute angular distance as
\begin{equation}
d(\theta_i, \theta_j) =
\mathrm{atan2}\big(\|u_i \times u_j\|,\,
\mathrm{clip}(u_i \cdot u_j,\,-1,\,1)\big),
\end{equation}
where the unit viewing direction for yaw $\psi$ and pitch $\phi$ is
\begin{equation}
u(\psi, \phi) =
[\cos\phi\cos\psi,\; \sin\phi,\; \cos\phi\sin\psi].
\end{equation}
We further apply a lightweight two-segment swap optimization to remove local discontinuities in the traversal order.

\noindent\textbf{Spatial Memory Selector.}

For each target view $\theta$, we select its $k$ nearest already processed neighbors within an angular threshold $\tau$ as memory references.
We set $k=6$ and $\tau=75^{\circ}$ in all experiments.
This Spatial Memory Selector restricts mask propagation to geometrically relevant views, rather than relying on adjacent indices in a fixed sequence.

Figure~\ref{fig:spsam2} illustrates the difference between temporal tracking and spatial memory selection.
The original SAM2 tracking mode uses nearby sequence indices as memory support, which can be suboptimal because sequence order does not necessarily match geometric proximity on the viewing sphere.
In contrast, Spatial SAM2 retrieves angularly nearby views, providing more relevant references for the target view and reducing accumulated mask errors on small structures such as wheels.

\noindent\textbf{Key frame initialization and filtering.}

Segmentation starts from the key view $\theta_0$, corresponding to the input SVG.
We obtain initial masks using automatic segmentation and apply lightweight filtering before propagation.
Specifically, we remove tiny regions, smooth local boundaries with a morphological closing operation, and suppress highly overlapping masks using non-maximum suppression.
The cleaned masks serve as stable prompts for Spatial SAM2 and initialize the multiview segmentation process.
A lightweight residual discovery step is applied after propagation to recover newly visible regions; details are provided in the supplementary material.

\subsection{Vector Refinement and Palette Alignment}
\label{sec:vector_domain_refinement}

Given the cross-view consistent masks, we crop each part into a transparent raster and vectorize it separately using VTracer~\cite{vtracer2023}.
This part-wise tracing reduces cross-part interference and prevents paths from crossing semantic boundaries.
The resulting part-level paths are then assembled into a single SVG for each view.

Raw vectorization can still introduce redundant micro paths and small color variations caused by antialiasing or raster rendering artifacts.
We therefore refine the generated SVGs in the vector domain.
For geometry, we remove stroke-only micro paths and merge redundant path fragments to improve compactness and editability.
For appearance, we align traced colors to the input SVG palette using the CIEDE2000 color-difference metric $\Delta E_{2000}$~\cite{sharma2005ciede2000}, with a small near-black bias to stabilize dark tones.
We also preserve alpha values and resolve inherited attributes into explicit colors for consistent rendering across SVG editors.
This refinement step reduces path redundancy, suppresses palette drift, and produces compact, editable, and cross-view consistent multiview SVG assets.

\section{Experiments}
\label{sec:exp}

\subsection{Implementation Details}
\label{sec:implementation}

We evaluate all multiview methods on the same benchmark of 200 input SVG examples.
For each example, we use the same 51-view yaw-pitch camera grid, matching the viewpoint setting available in Adobe Turntable.
Adobe Turntable outputs are manually exported from its interface, while \method{} and the related multiview baselines are evaluated on rendered views aligned to this grid whenever possible.

For the 3D generation stage, we employ the TRELLIS-image-large model to generate multiview rasterizations from the input SVG.
Our method and the open-source baselines are evaluated on a single NVIDIA A100-80GB PCIe GPU, while Adobe Turntable is evaluated through its cloud-based interface.

\subsection{Evaluation Metrics}
\label{sec:evaluation_metrics}

We evaluate the generated results from two perspectives: SVG asset stability and rendered multiview continuity.

\paragraph{SVG asset stability.}
For methods that output SVGs, we measure structural and color stability directly in the vector domain.
We report the path-count deviation $\mathrm{RMSE}_{\text{path}}$ across views, where lower values indicate more stable SVG structure.
We also report adjacent-view color-count variation $\overline{\Delta N}_{\text{color,nbr}}$ and perceptual color drift $\Delta E_{2000}$, where lower values indicate more stable palette behavior across viewpoints.
Runtime is measured for producing the full view set.

\paragraph{Rendered multiview continuity.}
To compare with methods that do not output SVGs, we render all results into images and evaluate local transition smoothness between adjacent views.
We compute DINOv2 feature consistency~\cite{oquab2023dinov2} and DreamSim perceptual similarity~\cite{fu2023dreamsim} separately along yaw and pitch transitions.
The overall score is computed as the average of the yaw and pitch scores.
Higher values indicate smoother and more consistent multiview transitions.

\subsection{Comparison}
\label{sec:comparison}

\begin{figure*}[htbp]
  \centering
  \includegraphics[width=0.9\linewidth]{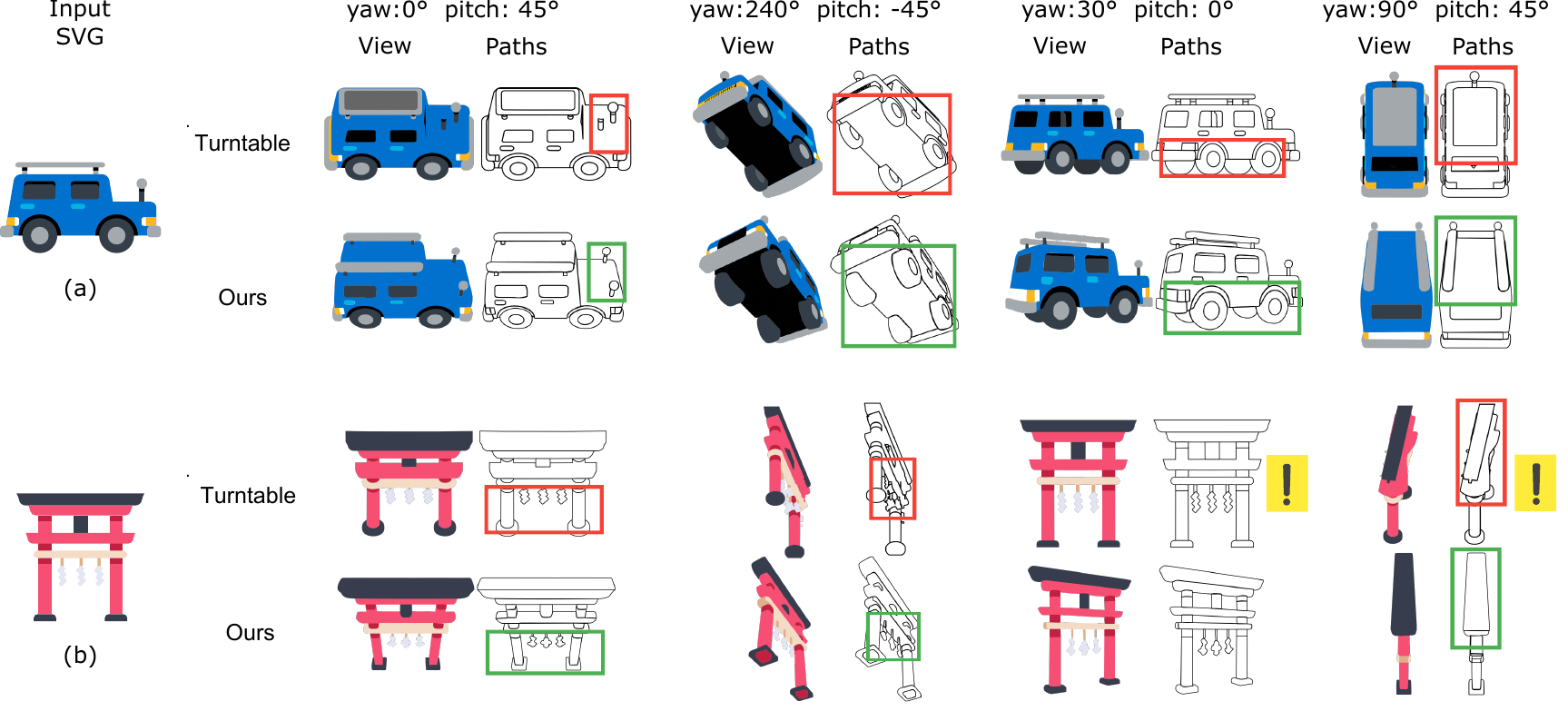}
  \caption{\textbf{Qualitative comparison with Adobe Turntable.}
  We compare Turntable and our method under the same target viewpoints, showing both rendered SVGs and vector path visualizations.
  In the vehicle example, Turntable produces inconsistent viewpoint geometry and unstable small parts: the front light changes or disappears across views, and the wheel-body perspective is less coherent.
  In the torii example, Turntable shows cluttered path structures and viewpoint-control failures.
  At yaw $30^{\circ}$, pitch $0^{\circ}$, its output remains close to the input pose instead of rotating to the target view; at yaw $90^{\circ}$, pitch $45^{\circ}$, the oblique side view is not correctly formed.
  Small structures such as the hanging ornaments under the crossbeam also become distorted or cluttered in the path visualization.
  In contrast, our method better preserves target viewpoints, small structures, and cleaner path topology across views.}
  \Description{A qualitative comparison figure with two examples, a blue vehicle and a torii gate. For each input SVG, several target viewpoints are shown. Each viewpoint compares Adobe Turntable and SVG360 using both rendered SVGs and vector path visualizations. Red boxes mark errors in Turntable outputs, such as inconsistent lights, distorted perspective, cluttered paths, missing small structures, and incorrect viewpoints. Green boxes mark the corresponding SVG360 results with more coherent geometry and cleaner paths.}
  \label{fig:comp_adobe}
\end{figure*}

\paragraph{Baselines.}
We compare \method{} with three groups of methods. 
First, Adobe Illustrator Turntable (Beta)~\cite{adobeTurntableHelp} is used as the direct SVG-oriented baseline, since it is one of the few publicly available tools that can generate multiple views from a single 2D vector artwork. 
For a broader comparison with related multiview generation systems, we also evaluate Wonder3D~\cite{long2024wonder3d} and Unique3D~\cite{wu2024unique3d}. 
These methods take the rasterized input image as input and generate multiview or 3D outputs, but they do not directly produce editable SVG path structures. 
We therefore include them only in rendered multiview continuity evaluation, not in SVG-specific structural metrics. 
Finally, for the single-view raster-to-SVG comparison, we compare with LIVE~\cite{yue2024live}, StarVector~\cite{wang2024starvector}, AutoTrace~\cite{weber2001autotrace}, and VTracer~\cite{vtracer2023}.

\noindent\textbf{Quantitative Comparison.}

\paragraph{Rendered multiview continuity.}
Table~\ref{tab:mv_continuity} compares \method{} with Adobe Turntable and related single-image multiview or 3D generation methods. 
\method{} achieves the highest DINO consistency on yaw, pitch, and overall scores, indicating stronger structural smoothness across adjacent viewpoints. 
It also obtains the best DreamSim pitch and overall scores. 
Turntable remains slightly higher on DreamSim yaw, which is consistent with its focus on horizontal turntable-style rotations, but \method{} performs more robustly on the full yaw-pitch grid. 
Compared with Wonder3D and Unique3D, \method{} provides stronger rendered continuity while additionally producing editable SVG outputs.

\begin{table}[htbp]
  \centering
    \caption{Rendered multiview continuity comparison.
    We measure local transition smoothness between adjacent views using DINOv2 feature consistency and DreamSim perceptual similarity.
    Overall scores are computed as the average of yaw and pitch scores.}
  \setlength{\tabcolsep}{2.5pt}
  \small
  \begin{tabular}{l ccc ccc}
    \toprule
    \multirow{2}{*}{Method}
    & \multicolumn{3}{c}{DINO Consistency}
    & \multicolumn{3}{c}{DreamSim Similarity} \\
    \cmidrule(lr){2-4}\cmidrule(lr){5-7}
    & Yaw$\uparrow$ & Pitch$\uparrow$ & Overall$\uparrow$
    & Yaw$\uparrow$ & Pitch$\uparrow$ & Overall$\uparrow$ \\
    \midrule
    Wonder3D
    & 0.8240 & 0.7925 & 0.8083
    & 0.8119 & 0.6890 & 0.7505 \\
    Unique3D
    & 0.8913 & 0.8467 & 0.8690
    & 0.8632 & 0.7773 & 0.8203 \\
    Turntable
    & 0.9028 & 0.8356 & 0.8692
    & \textbf{0.8892} & 0.7764 & 0.8328 \\
    Ours
    & \textbf{0.9171} & \textbf{0.8764} & \textbf{0.8968}
    & 0.8813 & \textbf{0.7904} & \textbf{0.8359} \\
    \bottomrule
  \end{tabular}
  \label{tab:mv_continuity}
\end{table}

\paragraph{SVG-centric stability.}
Table~\ref{tab:mv_svg_stats} compares \method{} with Adobe Turntable using SVG-specific metrics. 
\method{} reduces path-count deviation from 16.31 to 12.66, indicating more stable vector structure across viewpoints. 
It also lowers adjacent-view color-count variation from 2.31 to 0.43 and reduces perceptual color drift from 3.19 to 1.11, showing more stable palette behavior across views. 
For the full 51-view output set, \method{} takes 125 seconds on an A100 80GB GPU, while Turntable takes 187 seconds in its cloud-based interface. 
Although the runtime environments are not identical, the results suggest that \method{} is practical for generating editable multiview SVG assets.

\begin{table}[t]
  \centering
    \caption{SVG-centric comparison against Adobe Turntable.
    We report SVG structural stability, color consistency, and runtime for generating the full view set.
    Lower $\mathrm{RMSE}_{\text{path}}$, $\overline{\Delta N}_{\text{color,nbr}}$, and $\Delta E_{2000}$ indicate more stable structure and color across views.
    Runtime is reported under the available execution environment for each method.}
  \setlength{\tabcolsep}{4.5pt}
  \small
  \begin{tabular}{lcccc}
    \toprule
    Method
    & $\mathrm{RMSE}_{\text{path}}\downarrow$
    & $\overline{\Delta N}_{\text{color,nbr}}\downarrow$
    & $\Delta E_{2000}\downarrow$
    & Time (s)$\downarrow$ \\
    \midrule
    Turntable & 16.31 & 2.31 & 3.19 & 187 (cloud) \\
    Ours      & \textbf{12.66} & \textbf{0.43} & \textbf{1.11} & 125 (A100 80GB) \\
    \bottomrule
  \end{tabular}
  \label{tab:mv_svg_stats}
\end{table}

\paragraph{Single-view raster-to-SVG fidelity.}
Table~\ref{tab:single_view_svg} evaluates the initial-view vectorization quality by rendering each SVG back to a raster and comparing it with the input.
\method{} achieves the best SSIM, LPIPS, and $\Delta E_{2000}$ among LIVE, StarVector, AutoTrace, and VTracer, while maintaining a compact path count.
This verifies that our part-aware vectorization and SVG refinement preserve input appearance before extending the asset to multiple views.

\begin{table}[!t]
  \centering
  \caption{Single-view raster-to-SVG comparison on the initial input view. 
  We compare against LIVE~\cite{yue2024live}, StarVector~\cite{wang2024starvector}, 
  AutoTrace~\cite{weber2001autotrace}, and VTracer~\cite{vtracer2023}. 
  Predicted SVGs are rendered back to rasters and evaluated against the input raster.}
  \setlength{\tabcolsep}{5pt}
  \small
  \begin{tabular}{lcccc}
    \toprule
    Method & SSIM$\uparrow$ & LPIPS$\downarrow$ & $\Delta E_{2000}\downarrow$ & Paths$\downarrow$ \\
    \midrule
    LIVE       & 0.803 & 0.065 & 7.364  & 18.2 \\
    StarVector & 0.697 & 0.153 & 17.706 & \textbf{15.2} \\
    AutoTrace  & 0.812 & 0.063 & 2.792  & 448.2 \\
    VTracer    & 0.833 & 0.021 & 1.340  & 20.4 \\
    Ours       & \textbf{0.926} & \textbf{0.020} & \textbf{0.408} & 19.75 \\
    \bottomrule
  \end{tabular}
  \label{tab:single_view_svg}
\end{table}

\noindent\textbf{Qualitative comparison.}

\begin{figure*}[t]
  \centering
  \includegraphics[width=0.83\linewidth]{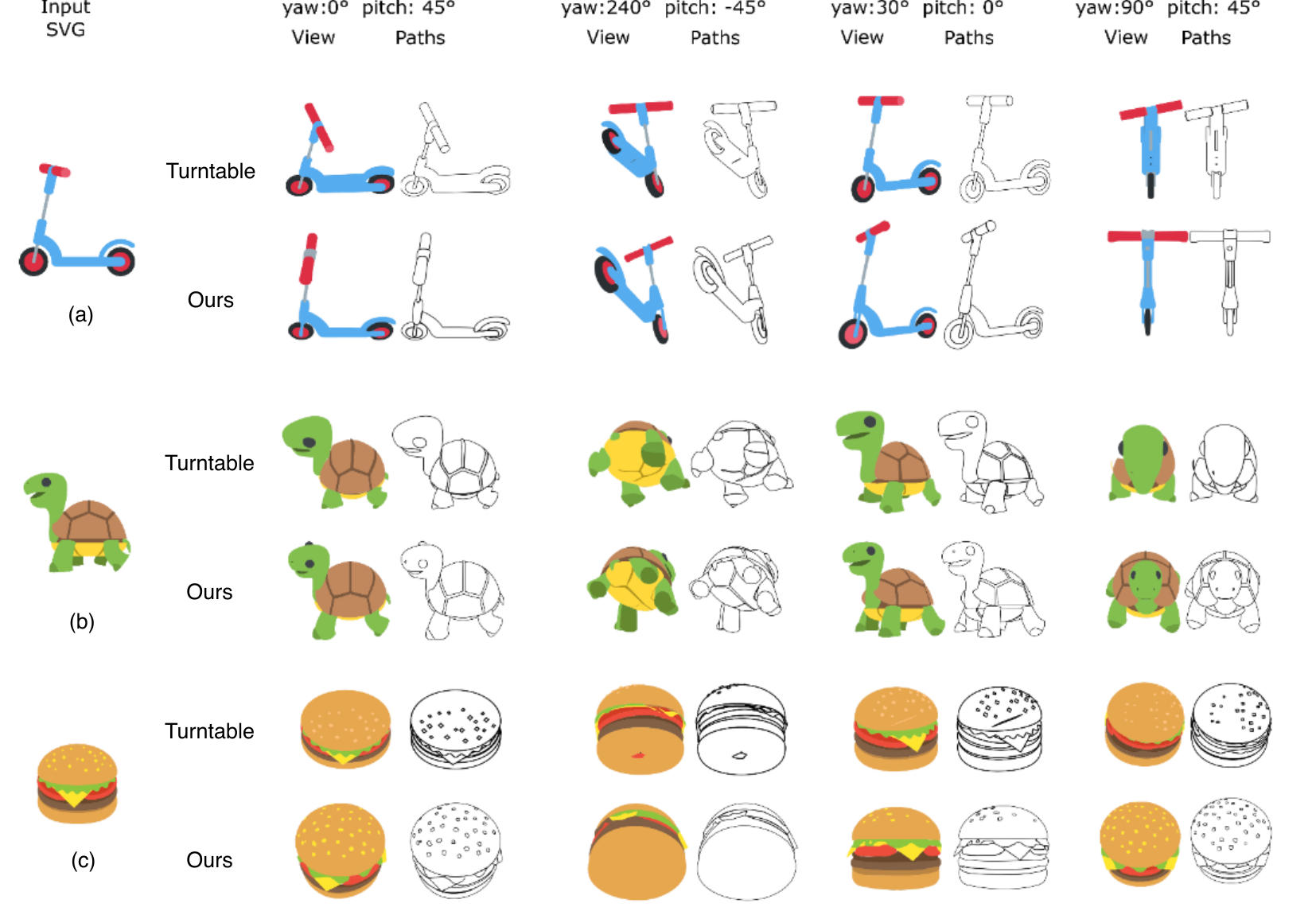}

  \caption{\textbf{Additional qualitative comparison with Adobe Turntable.}
  We compare Turntable and \method{} on the same target viewpoints, showing both rendered SVGs and vector path visualizations.
  The scooter, turtle, and hamburger examples reveal common Turntable artifacts, including geometric distortion, unstable outlines, cluttered path structures, and color or structure drift.
  \method{} produces cleaner path topology and more coherent multiview SVG assets across viewpoints.}
  \Description{A qualitative comparison figure showing scooter, turtle, and hamburger examples. Each row contains an input SVG and several target views. For each target view, Adobe Turntable and SVG360 are shown with both rendered SVG output and path visualization.}
  \label{fig:comp_adobe_extra}
\end{figure*}

\begin{figure*}[htbp]
    \centering
    \includegraphics[width=0.48\linewidth]{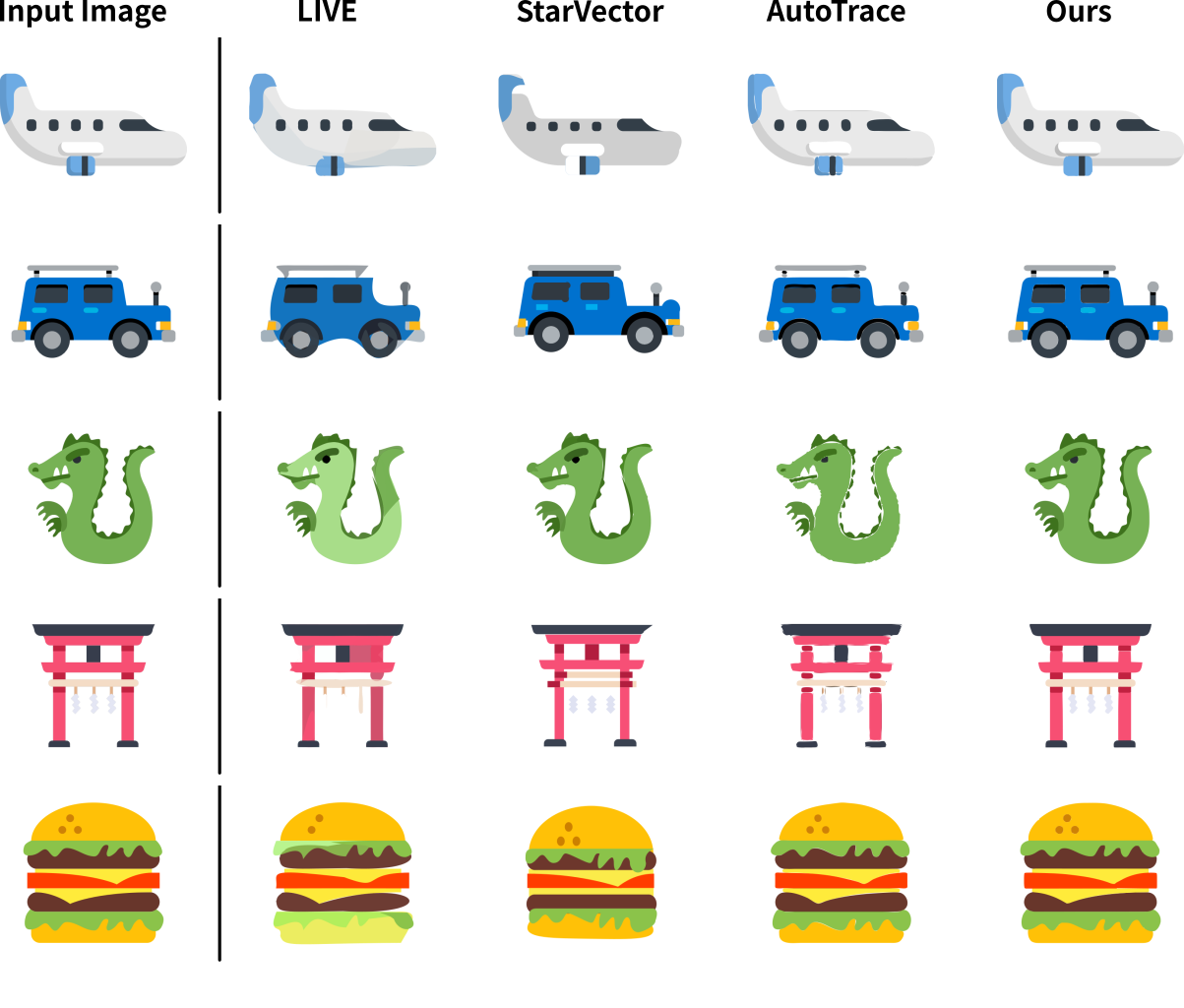} 
    \caption{\textbf{Qualitative single-view vectorization comparison.}
    Given the same input raster, we compare the rendered SVG outputs of different raster-to-SVG methods.
    Our method preserves cleaner object boundaries and more stable colors, while baselines often show blurry edges, color bleeding, missing details, or fragmented polygonal structures.}
    \Description{A qualitative comparison of single-view raster-to-SVG vectorization methods. The figure shows five example inputs, including an airplane, a vehicle, a dragon, a torii gate, and a hamburger. Each row compares the input raster with SVG outputs from LIVE, StarVector, AutoTrace, and the proposed method. The proposed method better preserves clean boundaries, colors, and object details.}
    \label{fig:supp_identical_view}
\end{figure*}

Figure~\ref{fig:comp_adobe} compares \method{} with Adobe Turntable.
Turntable often exhibits distorted geometry, cluttered paths, pose drift, or missing parts, whereas \method{} produces cleaner and more coherent multiview SVGs.
Additional Turntable comparisons and single-view vectorization results are shown in Figures~\ref{fig:comp_adobe_extra} and~\ref{fig:supp_identical_view}.

\subsection{Ablation Study}
\label{sec:ablation_study}

We ablate the role of spatially consistent segmentation by comparing four strategies: no SAM2 propagation, SAM2 Auto Mode, SAM2 Tracking Mode, and our Spatial SAM2. 
All variants use the same multiview raster inputs, part-wise vectorization, and SVG refinement stage, so the comparison isolates the effect of segmentation strategy on the final SVG outputs.

Table~\ref{tab:seg_ablation_svg} reports SVG-level structural stability, color consistency, and view-to-view continuity. 
Without SAM2 propagation, the vectorization stage often relies on coarse image-level regions, leading to unstable path structure and lower multiview consistency. 
SAM2 Auto Mode improves local segmentation quality but lacks cross-view correspondence, resulting in the highest path-count deviation. 
Tracking SAM2 benefits from memory propagation, but its temporal ordering is not always aligned with geometric proximity on the viewing sphere. 
Our Spatial SAM2 achieves the best results across all metrics, reducing $\mathrm{RMSE}_{\text{path}}$ to 9.45 and improving DINO and DreamSim consistency to 0.8921 and 0.8432, respectively.

\begin{figure}[!t]
  \centering
  \includegraphics[width=0.99\linewidth]{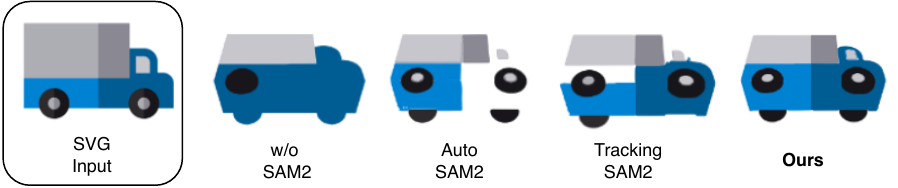}
  \caption{Qualitative ablation of segmentation strategies on final SVG outputs. 
  Without spatially consistent propagation, the generated SVGs exhibit missing parts, fragmented structures, or unstable topology. 
  Spatial SAM2 preserves more coherent part structure and produces cleaner multiview SVG results.}
  \Description{A qualitative ablation figure comparing an input SVG and four segmentation variants: without SAM2, SAM2 Auto Mode, SAM2 Tracking Mode, and Spatial SAM2. The figure shows that Spatial SAM2 produces the most coherent final SVG output, while other variants miss parts or produce unstable structures.}
  \label{fig:spsam4svg}
\end{figure}

Figure~\ref{fig:spsam4svg} provides a qualitative comparison on the final SVG outputs. 
Spatial SAM2 preserves more coherent part structure and produces cleaner multiview vector results, while the other variants show missing parts, over-fragmentation, or unstable topology.

\begin{table}[t]
  \centering
  \caption{Ablation of segmentation strategies on final multiview SVG outputs. 
  We evaluate structural stability, color consistency, and view-to-view continuity without requiring ground-truth part annotations.}
  \small
  \setlength{\tabcolsep}{4.5pt}
  \begin{tabular}{lcccc}
    \toprule
    Method
    & $\mathrm{RMSE}_{\text{path}}\downarrow$
    & $\Delta E_{2000}\downarrow$
    & \begin{tabular}{@{}c@{}}DINO\\Cons.$\uparrow$\end{tabular}
    & \begin{tabular}{@{}c@{}}DreamSim\\Sim.$\uparrow$\end{tabular} \\
    \midrule
    w/o SAM2       & 12.78 & 2.21 & 0.8144 & 0.7909 \\
    Auto SAM2      & 15.20 & 2.05 & 0.8388 & 0.8143 \\
    Tracking SAM2  & 13.74 & 1.97 & 0.8630 & 0.8285 \\
    Ours           & \textbf{9.45} & \textbf{1.03} & \textbf{0.8921} & \textbf{0.8432} \\
    \bottomrule
  \end{tabular}
  \label{tab:seg_ablation_svg}
\end{table}

\subsection{User Preference Study}
\label{sec:user_study}

\begin{figure}[t]
  \centering
  \includegraphics[width=\columnwidth]{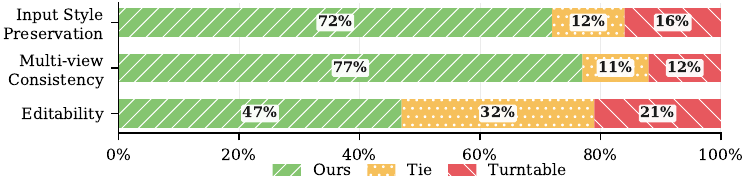}
  \caption{User preference study against Adobe Illustrator Turntable. 
  Participants choose \method{}, Turntable, or Tie under three criteria.
  \method{} is preferred across all criteria, especially for multiview consistency.}
  \Description{A horizontal stacked bar chart showing user preference percentages for input style preservation, multiview consistency, and editability. Each bar is divided into Ours, Tie, and Turntable.}
  \label{fig:user_study}
\end{figure}

We further conduct a paired user study against Adobe Illustrator Turntable, collecting 540 judgments from 15 participants over 12 examples.
For each comparison, participants view the input SVG and randomized outputs from Turntable and \method{}, then choose the preferred result or a tie under three criteria.
As shown in Figure~\ref{fig:user_study}, \method{} is preferred for input style preservation, multiview consistency, and editability.
The preference is strongest for multiview consistency, where 77\% of responses favor \method{}.
For editability, \method{} still receives the largest share of preferences, while the higher tie rate suggests that some outputs from both methods are perceived as similarly editable.

\subsection{Failure Cases}
\label{sec:failure_case}

\begin{figure}[t]
  \centering
  \includegraphics[width=\columnwidth]{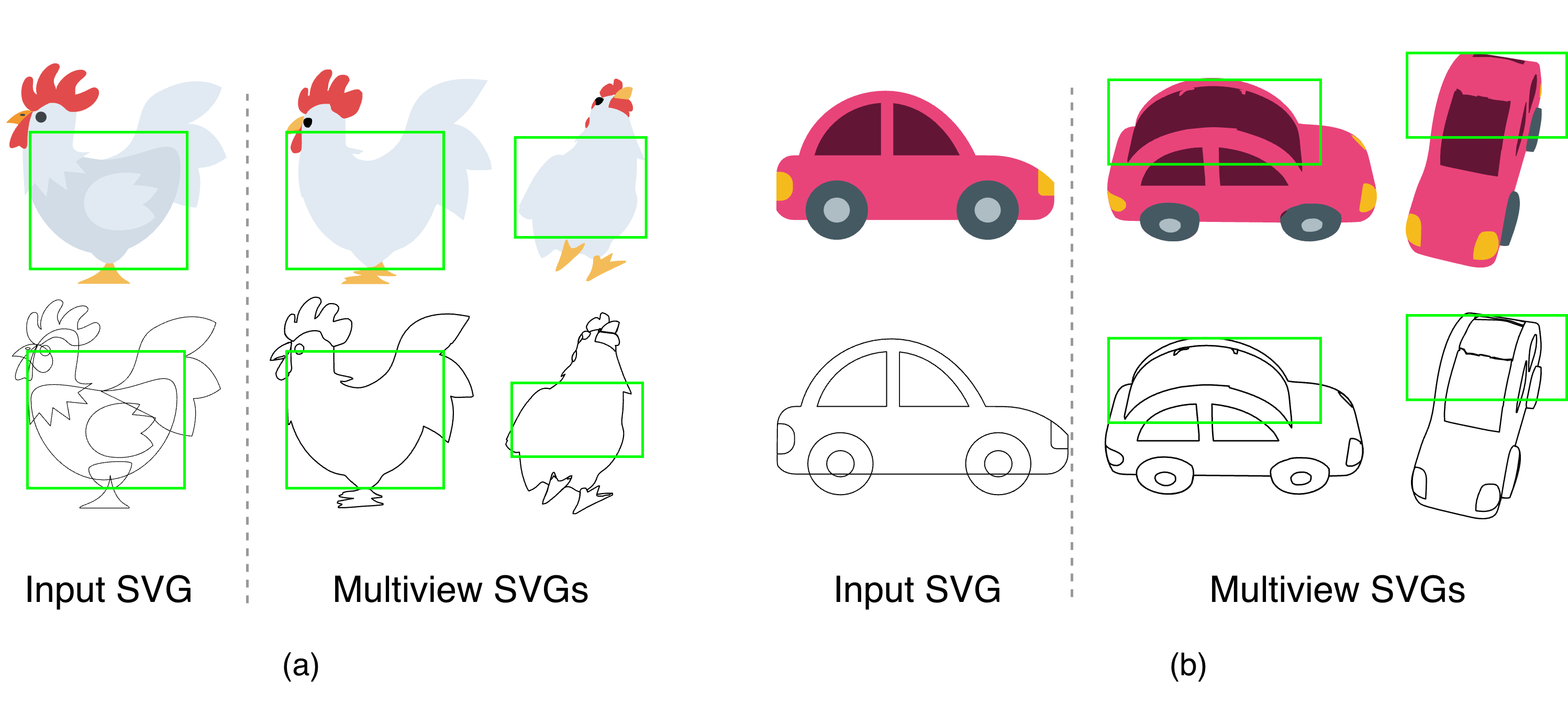}
    \caption{\textbf{Failure cases.}
    (a) Similar colors may lead to merged paths during refinement.
    (b) Small structures can still be missed under viewpoint changes, causing inconsistent part propagation.}
  \Description{A failure case figure with two examples. The first example shows a chicken where adjacent regions with similar colors are merged during refinement. The second example shows a car where a small roof window is present in one generated view but missing in another, indicating inconsistent part propagation.}
  \label{fig:failure_cases}
\end{figure}

Figure~\ref{fig:failure_cases} shows two representative failure modes. 
First, when neighboring parts have very similar colors, the refinement stage may merge them into a single region, reducing path-level separability, as shown in the chicken example. 
Second, Spatial SAM2 can still miss or inconsistently propagate small parts under large viewpoint changes or occlusions. 
In the car example, the roof window is preserved in one view but disappears in another. 
These cases suggest that future work should incorporate stronger cross-view part verification and more structure-aware color refinement.

\section{Conclusion}
\label{sec:conc}

We presented \method{}, a framework that converts a single input SVG into editable multiview SVG assets. 
By integrating 3D-aware multiview raster generation, spatially aligned part propagation, and SVG refinement with palette alignment, our method produces more coherent views, cleaner path structures, and stronger editability than Adobe Turntable and related baselines, as supported by quantitative results and user preferences.

Our method still has limitations. 
Similar neighboring colors may cause paths to merge during refinement, and small structures can be missed or inconsistently propagated under large viewpoint changes or occlusions. 
The current system also focuses on single-object SVGs and works best for closed, well-formed paths. 
Future work includes stronger cross-view part verification, better handling of open strokes and complex topology, scene-level vector generation, and more explicit use of vector structures such as layers, palettes, and path hierarchies.

%
%
%
%

\clearpage

\bibliographystyle{ACM-Reference-Format}
\bibliography{main}



\begin{figure*}[t]
  \centering
  \includegraphics[width=0.8\linewidth]{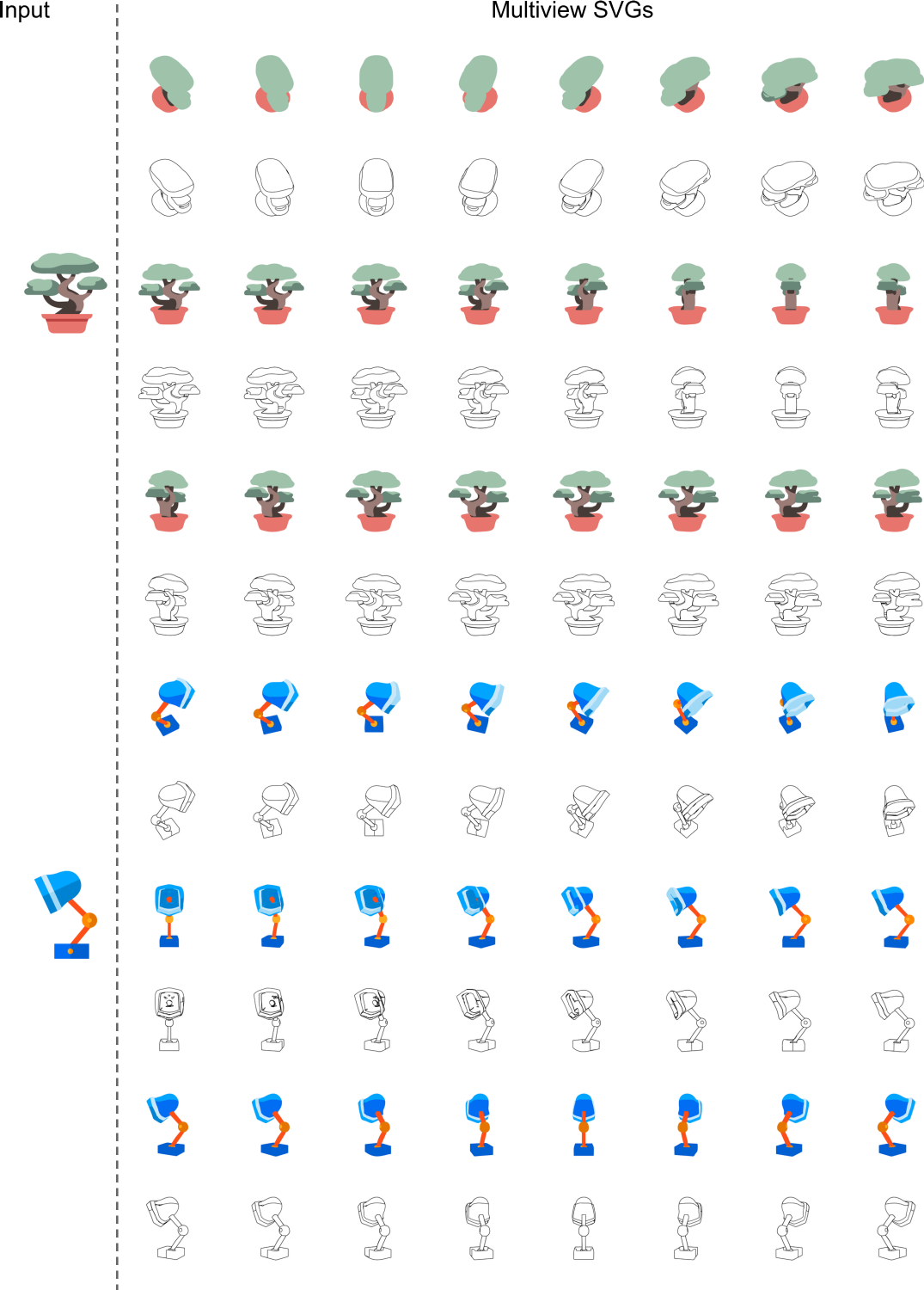}
  \caption{\textbf{Additional multiview SVG results.}
  For each example, we show the input SVG on the left and the generated multiview SVGs on the right.
  Each generated view is displayed together with its vector path visualization, illustrating that \method{} preserves the input style while producing editable SVG structures across viewpoints.
  The examples demonstrate consistent shape evolution, stable colors, and clean path topology under large viewpoint changes.}
  \Description{A qualitative result figure showing several input SVG examples and their generated multiview SVG outputs. The left column contains the input SVGs, while the right side shows multiple generated views for each example. Each colored SVG view is paired with a line-art vector path visualization. The figure demonstrates consistent appearance and editable path structure across viewpoints.}
  \label{fig:additional_results}
\end{figure*}



\end{document}